\documentclass[a4paper]{article}

\usepackage{INTERSPEECH2019}

\usepackage{lipsum}

\newcommand\blfootnote[1]{%
  \begingroup
  \renewcommand\thefootnote{}\footnote{#1}%
  \addtocounter{footnote}{-1}%
  \endgroup
}

\title{On Controlled DeEntanglement for Natural Language Processing }
\name{Sai Krishna Rallabandi}
\address{
  Language Technologies Institute, \\
  Carnegie Mellon University, \\
  Pittsburgh, PA 15213
  }
\email{srallaba@cs.cmu.edu}

\begin{document}

\maketitle

\begin{abstract}

Latest addition to the toolbox of human species is Artificial Intelligence(AI). Thus far, AI has made significant progress in low stake low risk scenarios such as playing Go and we are currently in a transition toward medium stake scenarios such as Visual Dialog. In my thesis\blfootnote{Presented at Doctoral Consortium, Interspeech 2019}, I argue that we need to incorporate controlled de-entanglement as first class object to succeed in this transition. I present mathematical analysis from information theory to show that employing stochasticity leads to controlled de-entanglement of relevant factors of variation at various levels. Based on this, I highlight results from initial experiments that depict efficacy of the proposed framework. I conclude this writeup by a roadmap of experiments that show the applicability of this framework to scalability, flexibility and interpretibility. 

\end{abstract}

\section{Motivation of Research}
\label{introduction}

In order to bring technology closer to humans it is important to provide mechanisms for them to interact with the same. Applications like Speech Recognition, Speech Synthesis, Visual Question Answering and Machine Translation to name a few have been shown to be massively useful in this context. Such Language Technologies are both rich in terms of quantity and diversity of tasks as well as inherent complexity due to the presence of human element. Hence they are the epitomy of medium risk medium stake applications forming a bridge between low stake applications such as playing games, spam filtering and high stake applications such as climate control, health related applications. Consequently, we can employ such applications as sanity tests in gauging the ability of AI - the latest addition to human toolbox - in fulfilling its potential to impact our lives. However, there are three broad problems that challenge this promise - (1) \textit{Scalability} - These technologies are currently only accessible in a handful number of languages around the planet. In order to have a meaningful impact it is imperative that such technologies need to at least exist in many more languages. (2) \textit{Flexibility} - Although deep learning based systems outperform their shallow learning counterparts in terms of quality, they still pale away in terms of flexibility and controllability. (3) \textit{Interpretibility} - Almost every deep learning system today is akin to a black box: we can neither interpret nor justify predictions by most of these models. In this thesis, I argue that employing a slightly different framework of learning - one with controlled de-entanglement of relevant factors of variation as a first class object - has the potential to address all the three issues simultaneously and henceforth rapidly accelerate progress in language technologies.  

\subsection{Controlled DeEntanglement}

Let us consider a typical deep learning architecture such as AlexNet\cite{alexnet}. It is characterized by a series of convolutional layers (feature extraction module) followed by a pooling layer and a SoftMax layer(classification module). Note that while I mention AlexNet as an example, this abstraction can be extended to most sequence to sequence architectures with encoder as feature extraction module and decoder as the classification module\cite{tutorial_dataaugmentation} across modalities and tasks. It can be shown that the pooling layer acts as information bottleneck\cite{tishby2000information} module in such architectures. I point out\cite{variational_attention_rsk} that in case of conventional Seq2Seq architectures deployed today, attention plays the role of information bottleneck module regulating the amount of information being utilized by the decoder. In \cite{variational_attention_rsk,vyas2019learning} I show that this module controls optimization in encoder decoder models leading to (1) Disentanglement of Causal Factors of variation in the data distribution (2) Marginalization of nuisance factors of variation from the input distribution. In case of models that employ stochasticity, two more effects can be observed : (a) Posterior collapse or Degeneration due to powerful decoders and (b) Loss of output fidelity due to finite capacity decoders. In current architectures, marginalization and disentanglement are realized implicitly and often lead to (a) and (b) when deployed in practise. I posit that designing learning paradigms such that we explicitly control de-entanglement of relevant factors of variation while marginalizing the nuisance factors of variation leads to massive improvements.  I refer to this as controlled DeEntanglement. Such an approach, I claim, leads to further advantages in the context of both generative processes - generation of novel content and discriminative processes - robustness of such models to noise and attacks.

\section{Key Issues identified and/or addressed}
The most popular approach to obtain disentanglement in neural models is by employing stochastic random variables. This approach provides flexibility to jointly train the latent representations as well as the downstream  network. It has been observed that the latent representations resemble disentangled representations under certain conditions \cite{isolating_sources_betavae, understanding_disentanglement_betavae,structured_disentangled_representations,hyperprior_disentanglement}. Note that although obtaining such degenerate representations is typical, it is not the only manifestation of disentanglement: it also manifests as continuous representations\cite{ravanelli2018interpretablesyncnet} and other abstract phenomena(e.g. grounding). I argue that explicitly controlling what and how much gets de-entangled \cite{understanding_disentanglement_betavae} is better than implicit disentanglement as is followed today\cite{locatello2018challenging}. I identify four ways to computationally control de-entanglement in encoder decoder models, by employing (1) suitable priors (2) additional adversarial or multi task objectives (3) an alternative formulation of probability density estimation and (4) a different objective of divergence. In my thesis, I present experimental findings from various tasks to show that the proposed framework has the ability to address all the three problems mentioned in section \ref{introduction}: \textit{scalability} via priors based controlled de-entanglement, \textit{flexibility} via priors, adversarial training and \textit{interpretibility} via alternative formulation of density estimation and objective of divergence.

\subsection{Scalability - Discovery of Acoustic Units and Styles}
\label{scalability}

A major bottleneck in the progress of many data-intensive language processing tasks such as speech recognition and synthesis is scalability to new languages and domains. Building such technologies for unwritten or under-resourced languages is often not feasible due to lack of annotated data or other expensive resources. I posit that there are two distinct categorizations that pose challenges in terms of scalability: (1) Unwritten languages and low resource scenarios and (2) Code switching and other non native speech phenomena. 

\subsubsection{Unwritten Languages}
Let us consider building speech technology for unwritten or under-resourced languages. A fundamental resource required to build speech technology stack in such languages is  phonetic lexicon: something that translates acoustic input to textual representation. Having such a lexicon - even if noisy and incomplete - can help bootstrap speech recognition and synthesis models which in turn enable other applications such as key word spotting. Therefore I proposed to employ controlled de-entanglement for unsupervised acoustic unit discovery in the context of our submission to ZeroSpeech Challenge 2019 \cite{cmu_zerospeech_2019}. We make an observation that articulatory information about speech production presents a discrete set of independent constraints. For instance, manner and place of articulation are two articulatory dimensions characterized by discrete sets(labial vs dental, etc). Based on this, we condition the prior space to conform to articulatory conditions by using a bank of discrete prior distributions. 

\subsubsection{Code switching and other non native Speech phenomena}

Speech has both continuous as well as discrete priors: The generative process of speech assumes a Gaussian prior distribution which is continuous in nature. However, the language which is also present in the utterance can be approximated to be sampled from a discrete prior distribution. Exact manifestation of this in linguistics can be at different levels: phonemes, words, syllables, sub word units, etc. Based on this insight, I show\cite{variational_attention_rsk} that incorporating priors help encode language independent information thereby facilitating synthesis of code mixed content. In addition to basing priors on knowledge about characteristics, I believe that it is also possible to base them on discovered patterns. In \cite{rallabandi_acoustic_styles}, we have discovered several code switching styles based on \cite{guzman2017metrics}. I plan to incorporate priors based on this style information to help speech recognition models targeted at decoding code switched speech.

\subsection{Flexibility - Global Control in Deep Generative Models}

We humans exhibit explicit global as well as fine grained control over how we deliver information. This enables us communicate more effectively in a conversation. The goal is to build AI models that can mimic this behavior. I have thus far worked on image captioning in the context of global control and emphatic text to speech in the context of fine grained control.

\subsubsection{Targeted Image Captioning}
In the context of image captioning, an interesting observation is that both the involved modalities - textual even though primarily symbolic and visual even though primarily spatial - are characterized by distinct discrete and continuous factors of variation. For instance, distinct objects or entities would intuitively perhaps be better represented by discrete variables, while their spatial location and relationships between them  might be represented by continuous variables.
Therefore, we split the latent prior space\cite{vyas2019learning} used for approximating the posterior distribution into continuous and discrete counterparts. Pressurizing the model to encode such prior information into the latent space provides us the flexibility to control the generative process by pinging different latent states during inference.

\subsubsection{Emphatic Text to Speech}
I am interested in investigating approaches to incorporate automatically derivable information from speech into the model architecture for better modeling and controlling prosody. 
In a sample implementation\footnote{Samples can be found here: https://t.ly/NExyy}, I have attempted disentangling heuristics about tonal information to accomplish local control. 
I first quantize fundamental frequency($F_0$) - a feature highly correlated with prosody - into multiple bins. I then incorporate the resultant discretized ordinal information at the phone level into our speech generation architecture in two different scenarios: (1) Explicit labeling scenario where we input the quantized $F_0$ values alongside the phoneme embeddings as input to the encoder and (2) Implicit incorporation as an additional task to predict these quantized $F_0$s at the output of encoder. The model was optimized using ordinal triplet divergence \cite{peter2019ordinal}. We show that our approach generates appropriate emphasis at word level and significantly outperforms AuToBI in terms of flexibility.

\subsection{Justification} \label{interpretbility}

Language is inherently composite in nature. Hence it is vital to model the abstract relationships so as to capture the unseen compositions of seen concepts at test time. However, accomplishing this is a deceptively non trivial task and might lead to models learning just surface level associations\cite{agrawal2017c,chen2017attacking,goyal2017making}. I present a case that flexible generative models provide additional information to improve performance in such tasks. Further, I hypothesize that when optimized using either a disjoint learning mechanism or a different divergence function, such models can also act as justifying modules for the task at hand. To ground this argument, I present two example applications that employ flexible models mentioned in the previous subsections: (1) Visual Question Answering System(VQA) that receivers additional information in the form of targeted captions. I propose to use Reward Augmented Maximum Likelihood\cite{RAML} to generate and integrate captions in the framework of Visual Question Answering. I show that tying the reward function to length of the generated caption forces the model to encode most relevant information thereby acting as justification to the selected answer. (2) Based on similar insights, I propose to apply\cite{RAML} to obtain and integrate speech recognition transcripts in the context of Acoustic Topic Identification System.

\section{Acknowledgements}

I sincerely thank Alan W Black for his invaluable guidance towards my dissertation. I thank AFRL for funding my research and student volunteers for taking part in the various subjective evaluations. I am very grateful to my collaborators for providing me an opportunity to work with them and the reviewers for useful feedback at various stages in my research career. 

\bibliographystyle{IEEEtran}
\bibliography{mybib}

\end{document}